\documentclass[letterpaper, 10 pt, journal, twoside]{IEEEtran}
\ifCLASSINFOpdf
\else
\fi
\usepackage{multicol}
\usepackage{multirow}
\usepackage{longtable}
\usepackage{threeparttable}
\usepackage{graphicx}
\usepackage{subcaption}
\usepackage{booktabs}
\usepackage{amsmath}
\usepackage{amssymb}
\usepackage{cite}
\usepackage{algorithm, algorithmic}
\usepackage{xcolor}
\usepackage{color}
\usepackage{amssymb}
\usepackage{pifont}

\usepackage{geometry}
\begin{document}
%
\title{ATI-CTLO: Adaptive Temporal Interval-based Continuous-Time LiDAR-Only Odometry}
%
%
%

\author{Bo Zhou, Jiajie Wu, Yan Pan and Chuanzhao Lu%
\thanks{Manuscript received: July 29, 2024; Revised: September 19, 2024; Accepted: October 7, 2024.}
\thanks{This paper was recommended for publication by Editor Javier Civera upon evaluation of the Associate Editor and Reviewers' comments.}
\thanks{This work was supported by the National Natural Science Foundation (NNSF) of China under Grant 620730075. (Corresponding author: Bo Zhou.)}
\thanks{Authors are with the School of Automation, Southeast University and the Key Laboratory of Measurement and Control of CSE, Ministry of Education, Nanjing 210096, P. R. China (email: zhoubo@seu.edu.cn; jiajiewu@seu.edu.cn; yanpan@seu.edu.cn;  lucz@seu.edu.cn)}
\thanks{Digital Object Identifier (DOI): see top of this page.}
}
%
%

\markboth{IEEE Robotics and Automation Letters. Preprint Version. Accepted October~2024}
{Zhou \MakeLowercase{\textit{et al.}}: ATI-CTLO} 

%



\maketitle

\begin{abstract}
The motion distortion in LiDAR scans caused by the robot's aggressive motion and environmental terrain features significantly impacts the positioning and mapping performance of 3D LiDAR odometry. Existing distortion correction solutions struggle to balance computational complexity and accuracy. In this letter, we propose an \textbf{A}daptive \textbf{T}emporal \textbf{I}nterval-based \textbf{C}ontinuous-\textbf{T}ime \textbf{L}iDAR-only \textbf{O}dometry (ATI-CTLO), which is based on straightforward and efficient linear interpolation. Our method can flexibly adjust the temporal intervals between control nodes according to the  motion dynamics and environmental degeneracy. This adaptability enhances performance across various motion states and improves the algorithms robustness in degenerate, particularly feature-sparse, environments. We validated our method's effectiveness on multiple datasets across different platforms, achieving comparable accuracy to state-of-the-art LiDAR-only odometry methods. Notably, in situations involving aggressive motion and sparse features, our method outperforms existing LiDAR-only methods.
\end{abstract}


%
\IEEEpeerreviewmaketitle

\section{Introduction}

\begin{figure}[t]
	\centering
	\includegraphics[width=0.48\textwidth]{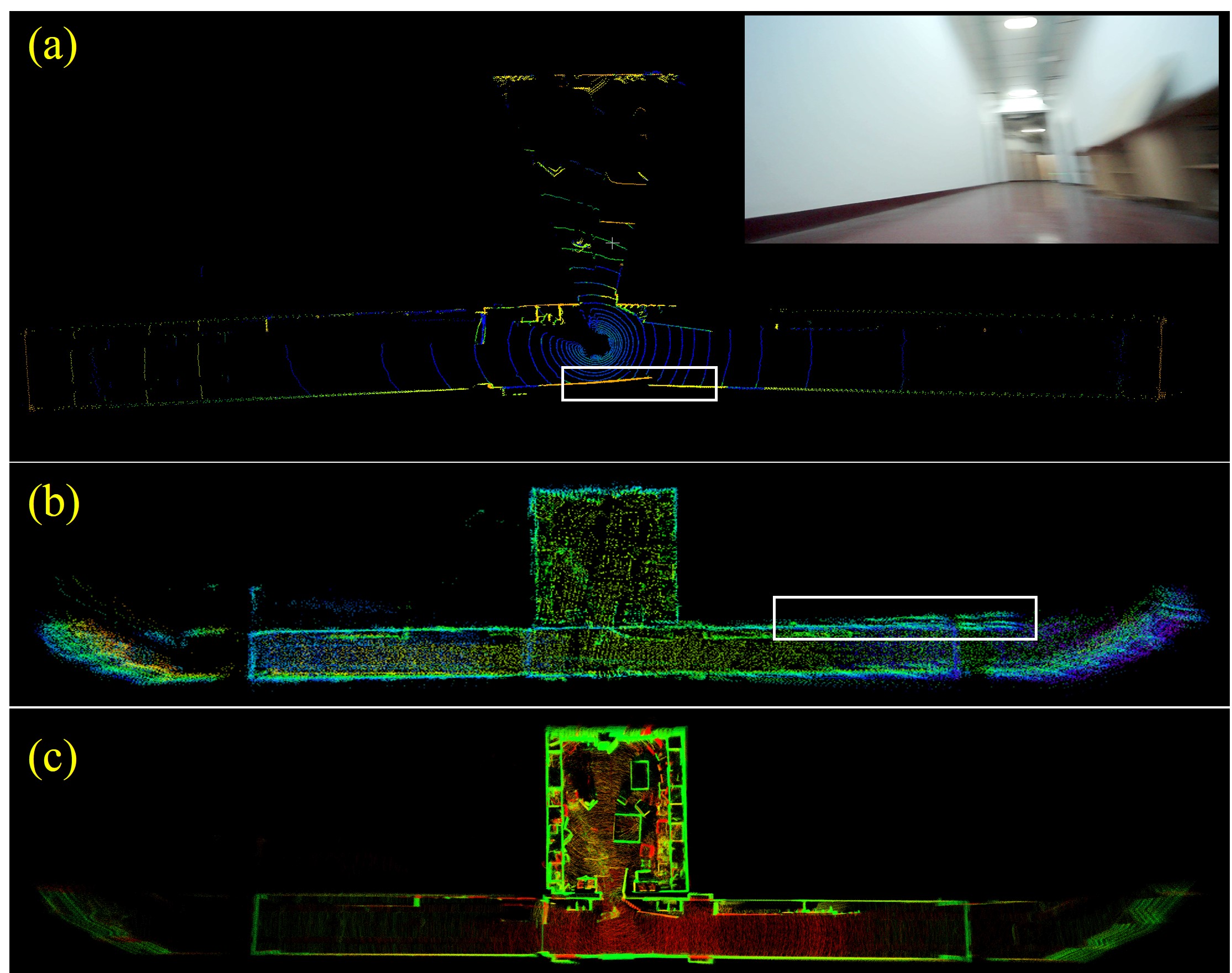}
	\caption{\small \textbf{Motion distortion correction}. \textbf{(a)}: A single frame of the point cloud is distorted due to aggressive turning. \textbf{(b)}: Correction distortion under constant velocity assumption (The mapping result of KISS-ICP~\cite{vizzo2023kiss}). \textbf{(c)}: Our method, utilizing dynamic temporal interval control, ensures more accurate mapping.}
	\label{fig:distorion_show}
	\vspace{-0.2in}
\end{figure}

\IEEEPARstart{L}{iDAR} odometry, recognized for its precise measurement capabilities and adaptability to varying light and weather conditions~\cite{lee2024lidar}, has been extensively utilized in robotics and autonomous driving, demonstrating remarkable progress over the last decade. Nonetheless, LiDAR odometry faces significant practical challenges, particularly the motion distortion in point clouds, which critically affects localization accuracy and map-building quality. There are several sources of motion distortion, including robot's movement during data acquisition, uneven terrain that causes the robot to oscillate, and environmental disturbances such as air currents affecting unmanned aerial vehicles (UAVs). Fig.~\ref{fig:distorion_show} illustrates the motion distortion resulting from aggressive turning during data acquisition and the mapping result of our method.

Most current LiDAR odometry methods are based on discrete-time methods, which assume LiDAR remains stationary during data acquisition and handle LiDAR distortion as part of preprocessing, aligning all points in the point cloud to a common reference frame based on constant velocity assumption~\cite{wang2021f,vizzo2023kiss}. This method, while straightforward and efficient, is inadequate for more aggressive motions (As Fig.~\ref{fig:distorion_show}b shows, ~\cite{vizzo2023kiss} remains distorted). In LiDAR-inertial odometry (LIO) systems, Inertial Measurement Units (IMUs) are used to directly measure the poses of LiDAR points to eliminate distortions~\cite{shan2020lio,xu2022fast}. However, this requires precise calibration and is highly susceptible to platform vibrations, such as mechanical vibrations of quadruped robots~\cite{yang2023cerberus}, and is limited by the IMU's measurement range in extreme motion conditions~\cite{he2023point}.

In contrast, continuous-time methods consider the LiDAR to be moving during data acquisition and model the trajectory as a function of time, allowing for pose queries at any given time, offering a novel approach to correcting motion distortion. 
Current models often assume constant velocity~\cite{dellenbach2022ct,nguyen2023slict,zheng2023ectlo} or higher-order modeling~\cite{droeschel2018efficient,quenzel2021real,chen2023direct}. However, higher-order models increase computational complexity, while lower-order models suffer from reduced precision. One solution to this problem relies on the assumption of motion continuity, using lower-order modeling but with smaller temporal intervals between the control nodes~\cite{dellenbach2022ct,zheng2024traj}. Smaller temporal interval allows the constant velocity motion prior to accommodate aggressive movements, but also introduces a new challenge. The small temporal intervals in continuous-time methods are often shorter than the duration of a single LiDAR frame, making it essential to segment the point cloud temporally. This segmentation reduces constraints between control nodes, leading to potential algorithm degeneracy or failure in feature-sparse environments.

To address the aforementioned challenges, we propose a continuous-time LiDAR-only method, the main contributions of our method are as follows:

\begin{itemize}
	\item We introduce \textbf{ATI-CTLO} using linear interpolation, allowing for flexible adjustment of temporal intervals between control nodes compared to other methods based on linear interpolation. The approach decreases the interval based on the dynamics of motion and increases the interval when the degeneracy occurs, effectively balancing accuracy and real-time performance.
	\item To address the issue of degeneracy due to cloud segmentation, we introduce a degeneracy management method for continuous-time LiDAR odometry. By categorizing the level of degeneracy based on the strength of environmental information constraints on the pose, we implement targeted treatments for each level. This approach significantly enhances the robustness of odometry in degenerate environments.
	\item Experiments conducted on various platforms, including unmanned ground vehicles (UGVs), UAVs, handheld devices, and quadruped robots, have demonstrated the effectiveness of our method. Under highly dynamic conditions, our algorithm achieves impressive improvement in accuracy compared to state-of-the-art (SOTA) methods. Additionally, in environments with sparse features, our method exhibits remarkable stability and outperforms all existing LiDAR-Only odometry methods.
\end{itemize}

\section{RELATED WORKS}
\subsection{Discrete-time and Continuous-time LiDAR Odometry}
3D LiDAR odometry technology primarily estimates pose through point cloud registration. The classic method, Iterative Closest Point (ICP) algorithm~\cite{besl1992method} is widely used. 
For instance, KISS-ICP~\cite{vizzo2023kiss} proposed a lightweight ICP method that dynamically adjusts the data association threshold, adapting to multiple scenarios with just one parameter file. 
Feature-based methods achieve rapid and precise registration by extracting significant points from the point cloud. LOAM~\cite{zhang2014loam} is a representative of this approach, inspiring subsequent advancements like LeGO-LOAM~\cite{shan2018lego}, which incorporated ground optimization, and F-LOAM~\cite{wang2021f}, which increased efficiency without sacrificing accuracy by adopting a scan-to-map method.

However, these methods rely on discrete-time representation, which inherently leads to LiDAR distortion issues. The common solutions involve using a constant velocity assumption or leveraging IMU pose estimation~\cite{shan2020lio,xu2022fast}. Continuous-time trajectory methods offer an alternative approach to removing motion distortions in point clouds, which allows query poses at any given time. Over the past decade, these techniques have made significant advancements in real-time performance and accuracy. Continuous-time methods can be divided into two main categories: nonparametric and parametric.

The nonparametric methods model the continuous-time trajectory as Gaussian Process (GP)~\cite{anderson2015full, dong2018sparse, le2019in2lama}, the GP prior smooths terms between discrete-time nodes. For parametric methods, B-spline is widely utilized to model the trajectory due to the locality and smoothness characteristics\cite{cioffi2022continuous}. Thanks to the efficient derivative computation of cumulative B-splines\cite{sommer2020efficient}, numerous high-performing B-spline-based continuous-time LiDAR odometry\cite{droeschel2018efficient,lv2021clins} approaches have been developed. These approaches excel in handling point cloud motion distortion and improving the smoothness of the trajectory. 

Besides these approaches, linear interpolation is also utilized for continuous trajectory modeling due to its efficiency and speed. CT-ICP~\cite{dellenbach2022ct} set poses at the start and end of each point cloud frame, using linear interpolation to fit the motion within the frame. However, it is not suitable for highly dynamic scenarios. Traj-LO~\cite{zheng2024traj} employed higher frequency pose controls to handle more aggressive movements, though this reduces the observational constraints between control nodes. To overcome the disadvantages of fixed control nodes, Coco-LIC~\cite{lang2023coco} integrated non-uniform B-splines, dynamically adjusting B-control nodes through IMU pre-estimation of motion. Nevertheless, this approach relies on IMU pre-estimation and suffers from reduced observational data in high-frequency interpolation scenarios.

\subsection{Degeneracy Detection}
LiDAR odometry is highly susceptible to degeneracy in environments with sparse or repetitive geometric features. Continuous-time LiDAR odometry, which uses smaller time intervals for more accurate trajectory fitting, exacerbates the risk of degeneracy due to the reduced number of LiDAR points between adjacent control nodes. Hence, degeneracy detection is essential for improving the robustness of LiDAR odometry.

Zhang et al.~\cite{zhang2016degeneracy} determined degeneracy by comparing the smallest eigenvalue $\lambda_{min}$ of the Hessian matrix to a predefined threshold, separating the degenerate directions in the state space and optimizing only in well-constrained directions. Hinduja et al.~\cite{hinduja2019degeneracy} applied this method to ICP, and using the ratio between the largest and smallest eigenvalues as the eigenvalue threshold. Ren et al.~\cite{ren2021lidar} suggested that pose updates in the degenerate direction cause oscillations and calculate the covariance matrix of pose updates during optimization, summing the diagonal elements as the degeneracy factor. Nubert et al.~\cite{nubert2022learning} introduced a learning-based degeneracy detection method, directly estimating localizability on raw LiDAR scans.

Another approach directly identifies the degenerate directions through the geometric characteristics of the environment. Yang et al.~\cite{yang2022recognition} utilized Principal Component Analysis (PCA) to analyze the normal vectors of all points in the point cloud, also using the smallest eigenvalue $\lambda_{min}$ as a criterion for describing degeneracy. Zhen et al.~\cite{zhen2019estimating} evaluated the strength of the constraint by measuring the sensitivity of measurements w.r.t the robot pose, but the method requires a prior-build map of the environment. X-ICP~\cite{tuna2023x} employed a similar method to~\cite{zhen2019estimating}, and evaluated the contribution of information pairs to the constraints. The method categorized localizability into three levels \{\textit{localizable, partially localizable, nonlocalizable}\} and enhanced constraints for partial localizability through resampling the available information.

\section{METHODOLOGY}

\subsection{System Overview}
\begin{figure}[t]
	\centering
	\includegraphics[width=1.0\linewidth]{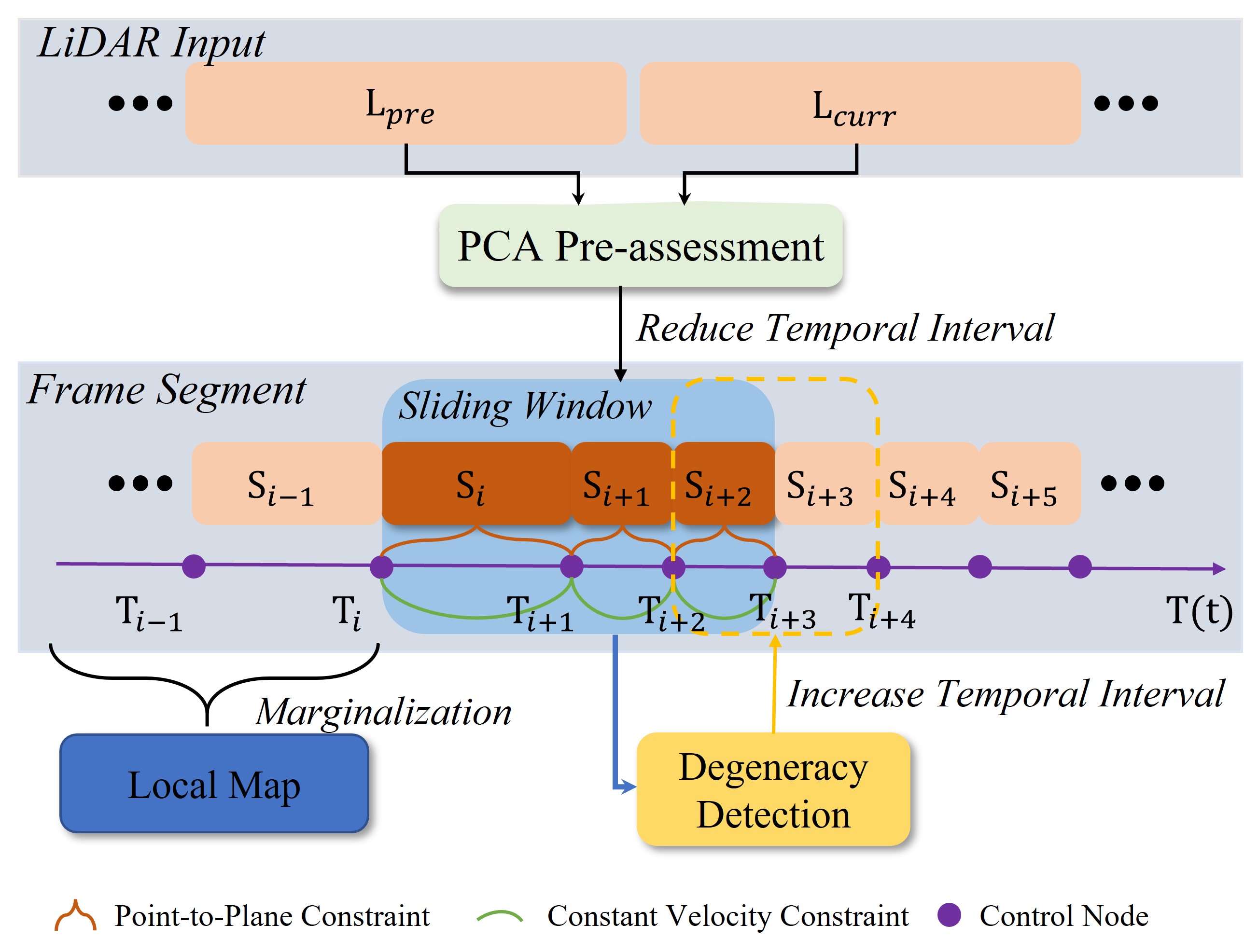}
	\caption{\small \textbf{System Overview}. Firstly, LiDAR scans segmentation is based on PCA pre-assessment, followed by solving through a sliding window approach (window's size $M$: 3). Simultaneously, degeneracy management merges cloud segments when degeneracy occurs. }
	\label{fig:pipeline}
	\vspace{-0.2in}
\end{figure}

The Fig.\ref{fig:pipeline} shows our pipeline. We denote the trajectory as $\mathbf{T}(t)$ in world frame, which is composed of continuous linear segments {$[\mathbf{T}_i,\mathbf{T}_{i+1}]$} over temporal interval {$[t_i, t_{i+1}]$}, the Lie Group {$\mathbf{T}_i \in \mathrm{SE(3)}$} represents the pose at timestamp $t_i$.  For the latest LiDAR input point cloud $\mathbf{L}_{curr}$, the {Principal Component Analysis (PCA)} technique is first used to evaluate the angle $\boldsymbol{\varphi}$ between the principal directions of $\mathbf{L}_{curr}$ and the previous point cloud $\mathbf{L}_{pre}$.
{Then the temporal interval $\Delta t$ between control nodes is reduced by a certain factor $\alpha$ ($\Delta t\leftarrow\alpha \Delta t$) once the angle $\boldsymbol{\varphi}$ exceeds the predefined threshold $\mathcal{K}_{vec}$ (Section \ref{subsec:PCA}).	
	After $\Delta t$ has been confirmed, the points in $\mathbf{L}_{curr}$ are divided into cloud segments $ \{\mathbf{S}_{i+1}, \mathbf{S}_{i+2}, \cdots , \mathbf{S}_{i+5} \}$ according to $\Delta t$, each cloud segment $\mathbf{S}_{i}$ corresponds to a continuous linear segment $[\mathbf{T}_i,\mathbf{T}_{i+1}]$. A sliding window approach is utilized to optimize multiple segments $\{\mathbf{S}_{i}, \mathbf{S}_{i+1},\cdots, \mathbf{S}_{i+M-1}\}$ at once ($M$ is the capacity of the sliding window).}
Once the optimization is finished, the sliding window {moves backward} to include new linear segments, while the oldest segment is added to the local map (Section \ref{subsec:odometry}). 
Notably, during the iterative optimization, the degeneracy detection module is applied to determine if the temporal interval should be increased, which is then achieved by merging linear segments (Section \ref{subsec:degradation}).

\subsection{PCA Pre-assessment to Reduce Temporal Interval }\label{subsec:PCA}

\begin{figure}[h]
	\begin{minipage}{1\linewidth}
		\centerline{\includegraphics[width=\textwidth]{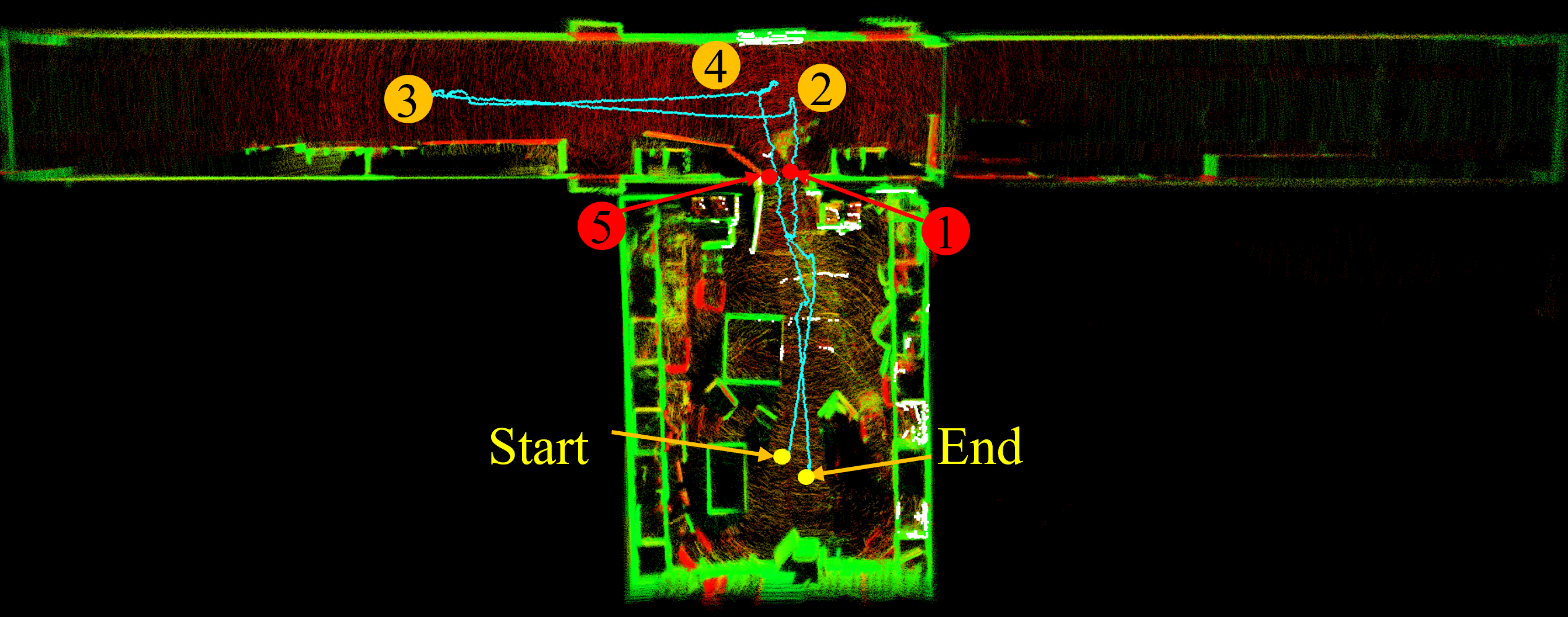}}
		\vspace{3pt}
		\centerline{\includegraphics[width=\textwidth]{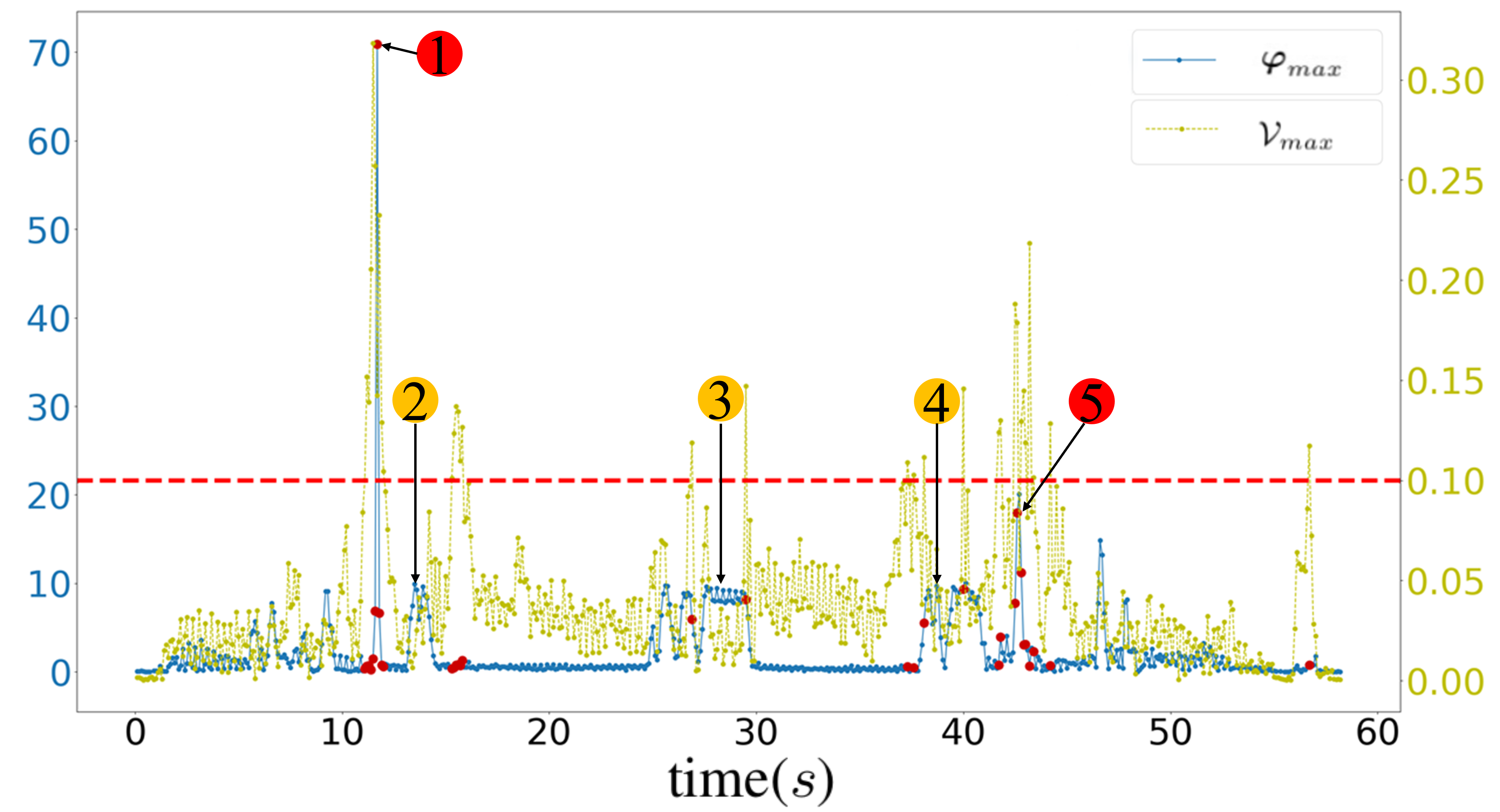}}
	\end{minipage}
	\caption{\small \textbf{PCA Pre-assessment}. A sudden change in the eigenvectors $\boldsymbol{v}_i$ indicates an aggressive turn in motion. In the \textbf{upper image}, the nodes with numbers show significant changes in the PCA direction. However, only the yellow nodes (2, 3, 4) are due to motion turns, while the red nodes (1, 5) result from environmental changes, such as the robot moving between a room and a hallway. The \textbf{line chart} below records the maximum PCA direction change $\boldsymbol{\varphi}_{max}$ ({blue}) and the maximum eigenvalue change $\mathcal{V}_{max}$ ({yellow}) throughout the entire trajectory. At nodes 1 and 5, there is a significant change in $\mathcal{V}_{max}$, and we filter out erroneous estimates (red nodes) by setting a threshold (red line).}
	\label{fig:pca_example}
	\vspace{-0.1in}
\end{figure}
The first-order linear motion model is straightforward and efficient but introduces significant errors during aggressive motions. To address this, higher-order motion models can be used, although they tend to be redundant and waste computational resources in stable conditions. Another approach involves using denser node configurations to handle aggressive motions. Therefore, predicting rapid motion changes is critical for effective node placement. Coco-LIC~\cite{lang2023coco} used IMU measurements for preliminary estimates, but for LiDAR-only odometry, the key challenge is to quickly estimate motion states based solely on LiDAR scans.

{In this letter, the PCA technique is utilized to quickly assess changes in motion states. As shown in Fig.~\ref{fig:pca_example}, during the robot's turning process (yellow nodes 2, 3, 4), the principal direction of the point cloud undergoes significant changes (blue). Note that we only concentrate on the changes in the principal direction caused by violent rotations, since constant velocity prior introduces significant errors when such rotations occur.}

{Considering all $N$ points  $^L\mathbf{p}_j$ in $\mathbf{L}_{curr}$, we first compute the covariance matrix $\mathbf{C} \in \mathbb{R}^{3\times3} $ of all points:}

\begin{equation}
	\label{equ: comput cov}
	\mathbf{C} = \sum_{j=1}^{N}(^L\mathbf{p}_j - \overline{\mathbf{p}})^{\top}(^L\mathbf{p}_j - \overline{\mathbf{p}}),
	\quad \overline{\mathbf{p}} =\frac{1}{N}\sum_{j=1}^{N}{^L\mathbf{p}_j}
\end{equation}

Then Singular Value Decomposition (SVD) is applied to $\mathbf{C}$ to derive the eigenvalues $\lambda_0 < \lambda_1 <\lambda_2$ and corresponding eigenvectors $\boldsymbol{v}_0, \boldsymbol{v}_1, \boldsymbol{v}_2$. The eigenvectors represent the principal directions of LiDAR scans, so the angles $\boldsymbol{\varphi} = [\boldsymbol{\varphi}_1, \boldsymbol{\varphi}_2,\boldsymbol{\varphi}_3]$ between the principal directions $\boldsymbol{v}^{curr}, \boldsymbol{v}^{last}$ of adjacent LiDAR scans indicate the dynamics of motion.
\begin{equation}
	\label{equ: calculate angle}
	\boldsymbol{\varphi}_i  = arccos(\frac{\boldsymbol{v}^{curr}_i \cdot \boldsymbol{v}^{last}_i}{|\boldsymbol{v}^{curr}_i||\boldsymbol{v}^{last}_i|}), i = 0,1,2
\end{equation}

However, the principal directions derived from PCA can abruptly change due to changes in the environment. During our experiments, we observed that these environmental changes also lead to variations in the eigenvalues, as illustrated in Fig.~\ref{fig:pca_example} (red nodes 1, 5). Leveraging this observation, we filter out changes in the PCA principal directions resulting from environmental changes:
\begin{subequations}
	\begin{align}
		\label{equ:compute v}
		\mathcal{V}_{max} &= max(\frac{\lambda^{curr}_i}{\lambda^{last}_i} - 1), i=0,1,2 \\
		\label{equ:compute final phi}
		\boldsymbol{\varphi}_{final} &= \left\{ 
		\begin{aligned}
			&\boldsymbol{\varphi},     &\mathcal{V}_{max} < \mathcal{K}_{val} \\
			&\boldsymbol{0}\in \mathbb{R}^{3\times1}, &otherwise \\
		\end{aligned}
		\right.
	\end{align}		
\end{subequations}

{Only when the max change in eigenvalues $\mathcal{V}_{max} < \mathcal{K}_{val}$ do we consider the principal direction change to be caused by motion. In this case,  we compare the maximum  change in the point cloud's principal direction $\boldsymbol{\varphi}_{max} = max(\boldsymbol{\varphi}_{final}) = max(\boldsymbol{\varphi}_1, \boldsymbol{\varphi}_2,\boldsymbol{\varphi}_3)$. If $\boldsymbol{\varphi}_{max} > \mathcal{K}_{vec}$, it indicates a significant change in the platform's movement, particularly in orientation, prompting a reduction in the point cloud segment size: $ \Delta t \leftarrow \alpha \Delta t$. Alternatively, if $\mathcal{V}_{max} \geq \mathcal{K}_{val}$, this suggests that the principal direction change is due to environmental changes. Under these circumstances, 
	$\boldsymbol{\varphi}_{final}$ is assigned a value of zero, and the point cloud segmentation is performed based on the initial temporal interval $\Delta t$.}

\subsection{Non-Uniform Linear Interpolation Continuous-time LiDAR-Only Odometry {Based} on Sliding Window} \label{subsec:odometry}
The continuous-time LiDAR odometry addresses the following problem: given the cloud segments $\mathbf{S}_i, i=0,1,2,\cdots$, estimate the corresponding linear segments $ [\mathbf{T}_i, \mathbf{T}_{i+1}]$ in the continuous trajectory $\mathbf{T}(t)$. Each linear segment $ [\mathbf{T}_i, \mathbf{T}_{i+1}]$ is solved through point-to-plane registration, constant velocity constraint.
In summary, we solve the following problem to get the control nodes $\mathbf{X} = (\mathbf{T}_i,\mathbf{T}_{i+1},\cdots,\mathbf{T}_{i+M}) $ in the sliding window, $M$ is the capacity of the sliding window:
\begin{equation}
	\label{equ:all_cost}
	\mathop{\text{arg\ min}}_{\mathbf{X}}{\mathcal{F}(\mathbf{X})}=\mathop{\text{arg\ min}}_{\mathbf{X}} \mathcal{F}_{ptpl}(\mathbf{X}) + \mathcal{F}_{vel}\left(\mathbf{X}\right) + \mathcal{F}_m(\mathbf{X})
\end{equation}
where the $\mathcal{F}_{ptpl}(\mathbf{X})$ is the point-to-plane constraint, $\mathcal{F}_{vel}(\mathbf{X})$ is the constant velocity constraint. $\mathcal{F}_m(\mathbf{X})$ is the marginalization energy that keeps information on the removed variables when the sliding window moves. {The error functions $\mathcal{F}_{ptpl}(\mathbf{X})$ and $\mathcal{F}_{vel}(\mathbf{X})$ of the whole sliding window consists of $M$ sub-error functions, which correspond to $M$ cloud segments $\{\mathbf{S}_{i}, \mathbf{S}_{i+1},\cdots, \mathbf{S}_{i+M-1}\}$:}
\begin{subequations}
	\begin{align}
		\label{equ:sub-cost-ptpl}
		\mathcal{F}_{ptpl}(\mathbf{X}) &= \sum_{k=0}^{M-1} {\mathcal{F}_{sub\_ptpl}}\left(\mathbf{T}_{i+k},\mathbf{T}_{i+k+1}\right) \\
		\label{equ:sub-cost-vel}
		\mathcal{F}_{vel}(\mathbf{X}) &= \sum_{k=0}^{M-1} {\mathcal{F}_{sub\_vel}}\left(\mathbf{T}_{i+k},\mathbf{T}_{i+k+1}\right)
	\end{align}
	
\end{subequations}

Every single sub-error function is associated with the linear segment $ [\mathbf{T}_{i+k}, \mathbf{T}_{i+k+1}]$ and the cloud segment $\mathbf{S}_{i+k}$. For the point-to-plane constraint, {as described in~\cite{dellenbach2022ct}} :
\begin{subequations}
	\begin{align}
		\label{equ: point-to-plane}
		{\mathcal{F}_{sub\_ptpl}}(\mathbf{T}_{i+k}, \mathbf{T}_{i+k+1}) = \mathbf{e}_{p}(i+k)\mathbf{Q}^{-1}_{p}\mathbf{e}_{p}(i+k) \\
		\label{equ:pp_error}
		\mathbf{e}_{p}(i+k) = \sum_{j=1}^{N_s} \mathbf{n}^T_{q_j}(\mathbf{T}_j  \cdot ^{L}\mathbf{p}_j - ^{W}\mathbf{q}_j) 
	\end{align}
	\label{equ:6}
\end{subequations} 

{where the $\mathcal{F}_{sub\_ptpl}$ is derived from the point-to-plane residual $e_p$. Due to the constant velocity prior, $\mathbf{T}_j$ is calculated via linear interpolation.} 

{As for $\mathcal{F}_{sub\_vel}(\mathbf{T}_{i+k},\mathbf{T}_{i+k+1})$, differing from fixed time interval methods, a scaling factor $\boldsymbol{\gamma}_{i+k}$ is incorporated into the velocity constraint calculations:}
\begin{subequations}
\begin{align}
	&{\mathcal{F}_{sub\_vel}}(\mathbf{T}_{i+k},\mathbf{T}_{i+k+1}) = \mathbf{e}^{\top}_v(i+k)\mathbf{Q}^{-1}_v\mathbf{e}_v(i+k) \\
	&\mathbf{e}_v(i+k) = \notag\\ 
	&\quad\quad \mathrm{Log}(\mathbf{T}_{i+k}^{-1}\mathbf{T}_{i+k+1}) - \boldsymbol{\gamma}_{i+k} \cdot\mathrm{Log}(\mathbf{T}_{i+k-1}^{-1}\mathbf{T}_{i+k}) \\
	&\boldsymbol{\gamma}_{i+k} = (t_{i+k+1} - t_{i+k}) / (t_{i+k}-t_{i+k-1})
\end{align}
\end{subequations}

{As the sliding window moves backward, the cloud segment $\mathbf{S}_i$ is removed and control node $\mathbf{T}_i$ is marginalized, while $\mathbf{T}_{i+1}$, which is still relevant to $\mathbf{S}_{i+1}$, is retained. Hence, $\mathbf{X}_m$ also includes $\mathbf{T}_{i+1}$ coming from $\mathbf{S}_i$, with this relationship represented by $\mathbf{X}_m = marg (\mathbf{X})$. The energy $\mathcal{F}_m(\mathbf{X})$ keeps information from $\mathbf{X}_m$, which is commonly stored as:}
\begin{subequations}
\begin{align}
	\label{equ:fm}
	\mathcal{F}_m(\mathbf{X}) = \frac{1}{2}(\mathbf{X}_m- \mathbf{X}_m^{0})^{\top}&H_m(\mathbf{X}_m- \mathbf{X}_m^{0}) + b^{\top}_m(\mathbf{X}_m-\mathbf{X}_m^0) \\ 
	\label{equ: jm}
	H_m = J_m^{\top}J_m,& \quad b_m = J_m^{\top}r_m
\end{align}
\end{subequations}

where the $H_m$,$b_m$ is marginalization priors, the $\mathbf{X}_m^0$ is the linearization point, which is applied for calculating the \textit{Schur complement} $\tilde{H}$ and corresponding vector $\tilde{b}$. From the $H_m$, $b_m$, we can derive $J_m$ and $r_m$ as Equ.~(\ref{equ: jm}), please refer to~\cite{zheng2024traj} for more details.

\begin{figure*}[t]
\centering
\includegraphics[width=0.98\textwidth]{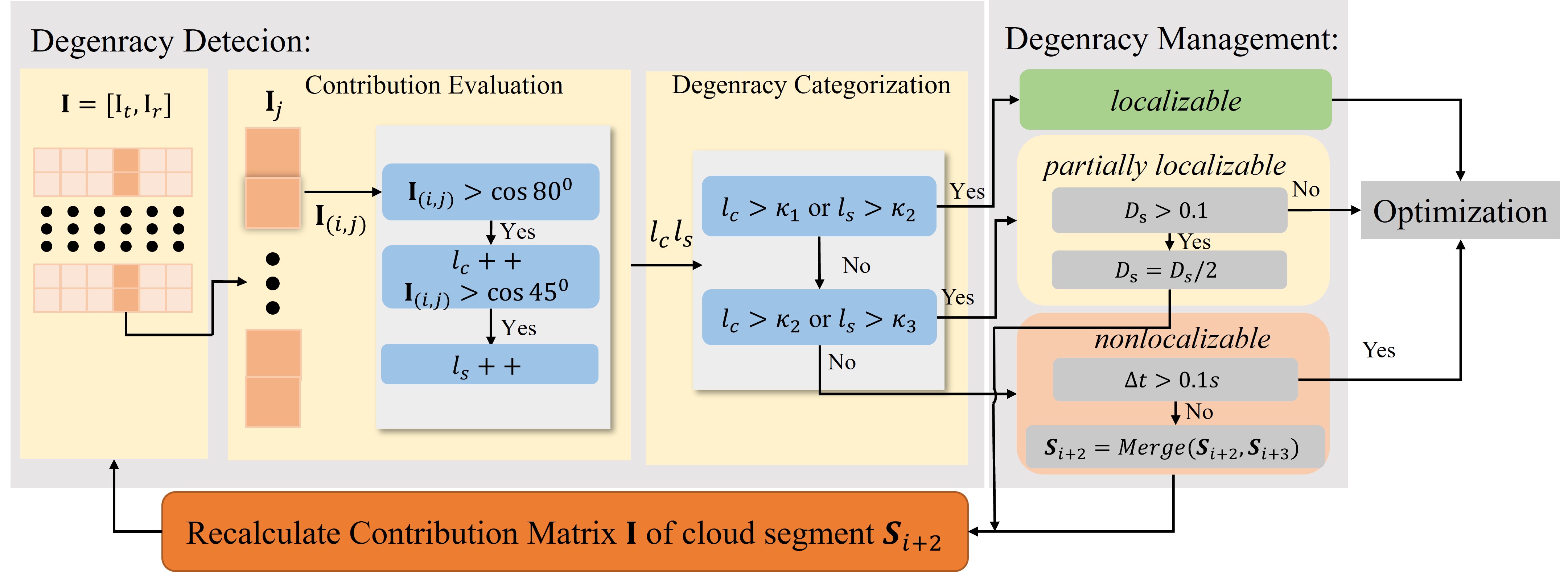}
\caption{\small \textbf{Degeneracy Detection:} Each column {$\mathbf{I}_j$} of the contribution matrix $\mathbf{I}$ represents the projection of environmental information in the corresponding principal direction. In the contribution evaluation, we calculate the contribution of each column {$\mathbf{I}_j$} and use $l_c$ and $l_s$ to count the number of information elements $\mathbf{I}_{(i,j)}$ with varying levels of contribution. Then different thresholds $\kappa_1, \kappa_2, \kappa_3 $ are used to classify the contributions into various levels, eventually the localizability of each direction is divided into three levels: \{\textit{localizable, partially localizable, nonlocalizable}\}. \textbf{Degeneracy Management:}  We enhance the number of effective information elements by increasing the sampling rate (decreasing $D_s$) for partially localizable cases. In nonlocalizable cases, subsequent cloud segment {$\mathbf{S}_{i+3}$} is merged to gather more valid information. }
\label{fig:degeneracy_management}
\vspace{-0.2in}
\end{figure*}

\subsection{Degeneracy Management to Increase Temporal Interval}\label{subsec:degradation}
\subsubsection{Degeneracy Detection}
As the linear segments $\mathbf{S}_i$ are obtained by slicing the LiDAR scans $\mathbf{L}$, they accumulate fewer LiDAR points compared to discrete-time methods, which increases the likelihood of degeneracy during optimization. 
In this letter, a classification of degeneracy levels is conducted using geometric feature-based methods, with specific handling approaches applied to each level. 
{As described in \cite{tuna2023x}\cite{chang2023wicrf2}, this approach evaluates the contribution of an information pair $(^L\mathbf{p}, ^L\mathbf{n})$ to the localizability in the principal directions of optimization by calculating the normal vector $^L\mathbf{n}$ and torque $\boldsymbol{\tau}  = ^L\mathbf{p} \times ^L\mathbf{n}$. 
So the information matrix is defined as: }

\begin{equation}
	\mathbf{F}_r = \begin{bmatrix}
		\frac{\boldsymbol{\tau}_1}{\left\| \boldsymbol{\tau}_1 \right\|_2} & \cdots  & 	\frac{\boldsymbol{\tau}_n}{\left\| \boldsymbol{\tau}_n \right\|_2}
	\end{bmatrix}^{\top}, 
	\mathbf{F}_t = \begin{bmatrix}
		^L\mathbf{n}_1	& \cdots & ^L \mathbf{n}_n
	\end{bmatrix}^{\top}
	\end{equation}
	
	{Differing from \cite{tuna2023x}, the sliding window is used to optimize several multiple control nodes at once, with each cloud segment $\mathbf{S}_i$ corresponding to two control nodes $T_i$ and $T_{i+1}$. Performing degeneracy detection on all variables within the sliding window would significantly reduce the algorithms real-time performance. In practice, only the latest point cloud segment within the sliding window is considered during implementation. For example, in Fig~\ref{fig:pipeline}, only the information from $\mathbf{S}_{i+2}$ is projected onto the optimization principal directions $\mathbf{V}_r$ and $\mathbf{V}_t$ corresponding to $\mathbf{T}_{i+3}$ to obtain the contribution matrix:}
	\begin{equation}
\mathbf{I}_r = (\mathbf{F}_r \cdot \mathbf{V}_r)^{\left| \cdot \right|} \in \mathbb{R}^{n\times3}, \quad \mathbf{I}_t = (\mathbf{F}_t \cdot \mathbf{V}_t)^{\left| \cdot \right|} \in \mathbb{R}^{n\times3}
\end{equation}
The $\left| \cdot \right|$ means getting the absolute value of all elements in the matrix. In matrix $\mathbf{I}=[\mathbf{I}_t, \mathbf{I}_r]$ indicates the contribution level of information pair $(^L\mathbf{p}_i, ^L\mathbf{n}_i)$ to direction $\mathbf{v}_{r_j}$ or  $\mathbf{v}_{t_j}$. We then divide the localizability of each direction into three levels by evaluating the contribution of information matrix, more details can be seen in Fig.~\ref{fig:degeneracy_management}.

\subsubsection{Degeneracy Management}
In discrete-time methods, the nonloclizable directions are excluded from the optimization, and prior motion values are used instead. For partially localizable directions, the constraints are calculated by resampling the available information instead.

In this letter,  degeneracy may arise from the segmentation of the point cloud, which reduces the number of constraint points. Thus, we must first exclude degeneracy caused by this factor. As {Fig.~\ref{fig:pipeline}} shows, if a particular direction is nonlocalizable, we will merge the {$\mathbf{S}_{i+2}$} with the next segment {$\mathbf{S}_{i+3}$} and then perform degeneracy detection again. To prevent unlimited merging, we stop the process when the temporal interval $\Delta t$ of segment {$\mathbf{S}_{i+2}$} exceeds $0.1s$. If a direction is partially localizable, it indicates that {$\mathbf{S}_{i+2}$} contains useful information. We then halve the voxel size for downsampling {$\mathbf{S}_{i+2}$} to obtain more available information pairs. Similarly, when the voxel size falls below 0.1, we move directly to the optimization.

\begin{table*}[!ht]
\centering
\caption{{ RMSE of ATE (m) / {RPE (m)} on M2DGR Dataset}}
\resizebox{0.98\textwidth}{!}{
	\begin{threeparttable}
		\begin{tabular}{ c c c c c c c c c c }
			\toprule
			\midrule
			\multirow{2}{*}{Approach}  & room\_01 & room\_dark\_06 & gate\_01 & hall\_02 & door\_02 & circle\_01 &street\_03 & street\_07 & \multirow{2}{*}{Average} \\
			& (27.1m) & (72.5m) & (139.2m) & (109.4m) & (57.3m) & (316.5m) & (423.9m) & (1104.6m) & \\
			\midrule
			LeGO-LOAM   & 0.136 / 0.203 & 0.305 / 0.217 & 0.328 / 0.482 & \textbf{0.239} / 2.693 & 0.187 / 2.364 & 1.529 / 0.710 & 0.597 / 0.802 & 10.584 / 0.169 & 1.738 / 0.955 \\
			
			KISS-ICP  & 0.219 / 0.114 & 0.339 / 0.125 & \textbf{0.141} / 0.147 & 0.276 / 0.809 & 0.263 / 0.406 & 1.499 / 0.254 & \textbf{0.131} / 0.256 & 0.400 / 0.164 & 0.408 / 0.284 \\
			
			CT-ICP  & {0.144} / 0.089 & 0.293 / 0.091 & 0.159 / 0.152 & 0.259 / 0.808 & 0.211 / 0.404 & 0.357 / 0.252 & 0.174 / 0.259 & \ding{55} / \ding{55} &  0.228 / 0.294 \\
			
			Traj-LO   & 0.148 / 0.043 & 0.293 / \textbf{0.036} & 0.151 / \textbf{0.056} & 0.257 / \textbf{0.362} & 0.196 / 0.170 & 0.357 / 0.103 & 0.173 / 0.126 & 0.249 / 0.065 & 	0.228 / 0.120 \\
			
			\textbf{Ours} & \textbf{0.132} / \textbf{0.040} & \textbf{0.284} / 0.037 & 0.151 / 0.059 & 0.251 / 0.364 & \textbf{0.185} / \textbf{0.157} & \textbf{0.354} / \textbf{0.099} & 0.170 / \textbf{0.101} & \textbf{0.212} / \textbf{0.064} & {\textbf{0.217}} / \textbf{0.115}\\
			
			\bottomrule
		\end{tabular}
		\begin{tablenotes}
			\item[1] \small{{'\ding{55}' denotes significant error, occurring when the estimated distance or orientation between consecutive control nodes exceeding $5m$ or $30^o$ \cite{dellenbach2022ct}. \qquad '(*m)' denotes the length of the sequence.}}
		\end{tablenotes}			
	\end{threeparttable}
}
\label{tab:m2dgr}
\vspace{-0.1in}
\end{table*}

\begin{table*}[t]
\centering
\caption{{RMSE of ATE (m) on NTU VIRAL Dataset}}
\resizebox{0.98\textwidth}{!}{
	\begin{threeparttable}
		\begin{tabular}{c c c c c c c c c c c c c c c c c c c c c c c}   
			\toprule
			\midrule
			\multirow{2}{*}{Approach} & \multirow{2}{*}{Sensor} & \multicolumn{3}{c}{eee} & \multicolumn{3}{c}{nya} & \multicolumn{3}{c}{sbs} & \multicolumn{3}{c}{rtp} & \multicolumn{3}{c}{tnp} & \multicolumn{3}{c}{spms}\\
			\cmidrule(r){3-5} 
			\cmidrule(r){6-8}
			\cmidrule(r){9-11}
			\cmidrule(r){12-14}
			\cmidrule(r){15-17}
			\cmidrule(r){18-20}
			& & 01 & 02 & 03 & 01 & 02 & 03 & 01 & 02 & 03 & 01 & 02 & 03 & 01 & 02 & 03 & 01 & 02 & 03\\
			& & (237.0m) & (171.1m) & (127.8m) & (160.3m) & (249.1m) & (315.5m) & (202.3m) & (183.5m) & (198.5m) & (321.2m) & (200.0m) & (304.5m) & (252.3m) & (139.6m) & (123.8m) & (545.0m) & (697.9m) & (312.4m)\\
			\midrule
			F-LOAM & L1 & 4.486 & 8.328 & 1.133 & 1.447 & 1.292 & 1.498 & 0.976 & 0.2010 & 1.079 & 10.775 & 4.637 & 2.218 & 2.354 & 2.249 & 1.566 & \ding{55} & \ding{55} & \ding{55}\\
			KISS-ICP & L1 & 2.220 & 1.570 & 1.014 & 0.628 & 1.500 & 1.272 & 0.917 & 1.312 & 1.030 & 3.663 & 1.970 & 2.382 & 2.305 & 2.405 & 0.799 & 8.493 & \ding{55} & 5.454 \\
			CT-ICP & L1 & 7.763 & 0.125 & 11.171 & 0.100 & 0.101 & 0.073 & \ding{55} & 0.084 & 1.545 & \ding{55} & 0.081 & 0.086 & \textbf{0.073} & \textbf{0.071} & \textbf{0.045} & \ding{55} & \ding{55} & \ding{55} \\
			Traj-LO & L1 & 0.055 & 0.039 & 0.035 & 0.047 & 0.052 & 0.050 & 0.048 & 0.039 & 0.039 &\textbf{0.050} & \textbf{0.058} & 0.057 & 0.505 & 0.607 & 0.101 & 0.121 & \ding{55} & 0.103 \\
			\textbf{Ours} & L1 & \textbf{0.046} & \textbf{0.033} & \textbf{0.031} & \textbf{0.044} & \textbf{0.048} & \textbf{0.047} & \textbf{0.032} & \textbf{0.030} & \textbf{0.032} & 0.054 & \textbf{0.058} & \textbf{0.056} & 0.378 & 0.360 & 0.091 & \textbf{0.072} & \textbf{0.077} & \textbf{0.079} \\

			\bottomrule
		\end{tabular}
		\begin{tablenotes}
			\small {\item[1] {'\ding{55}' denotes significant error, as defined in the same way as in Table~\ref{tab:m2dgr}. \qquad '(*m)' denotes the length of the sequence.}}
		\end{tablenotes}
	\end{threeparttable}
}
\label{tab:ntu_dataset}
\vspace{-0.2in}
\end{table*}

\section{EXPERIMENTS}
We validated the effectiveness of the ATI-CTLO algorithm on datasets from various platforms, including the UGV dataset M2DGR~\cite{yin2021m2dgr} , the UAV dataset NTU VIRAL~\cite{nguyen2022ntu} , {the handheld dataset Newer College Dataset ~\cite{ramezani2020newer}} and our own quadruped robot dataset. In comparisons with the SOTA LiDAR odometry methods, including discrete-time methods such as LeGO-LOAM~\cite{shan2018lego} and KISS-ICP~\cite{vizzo2023kiss}, and fixed-interval continuous-time odometry algorithms like CT-ICP~\cite{dellenbach2022ct} and Traj-LO~\cite{zheng2024traj}, our method ATI-CTLO demonstrated an ability to handle more challenging environments.

In this letter, we maintained the same parameters setting for all experiments. In the PCA pre-assessment module, when the change in eigenvalue is below $\mathcal{K}_{val} = 0.1$ and the angle $\boldsymbol{\varphi}_{max}$ exceeds $\mathcal{K}_{vec} = 10$, we halve the temporal interval ($\alpha = 0.5$). In the degeneracy management module, we used $\kappa_1 = 125, \kappa_2 = 50, \kappa_3 = 15$ to classify the levels of degeneracy. {For the optimization process, the initial temporal interval $\Delta t$ was set to $0.04s$, and the capacity of sliding window $M$ was set to $3$.} 
All experiments were conducted on an Intel i7-12700H CPU.

\subsection{Dataset Results}
\subsubsection{M2DGR Dataset}
\begin{figure}[h]
	\centering	
	\includegraphics[width=1.0\linewidth]{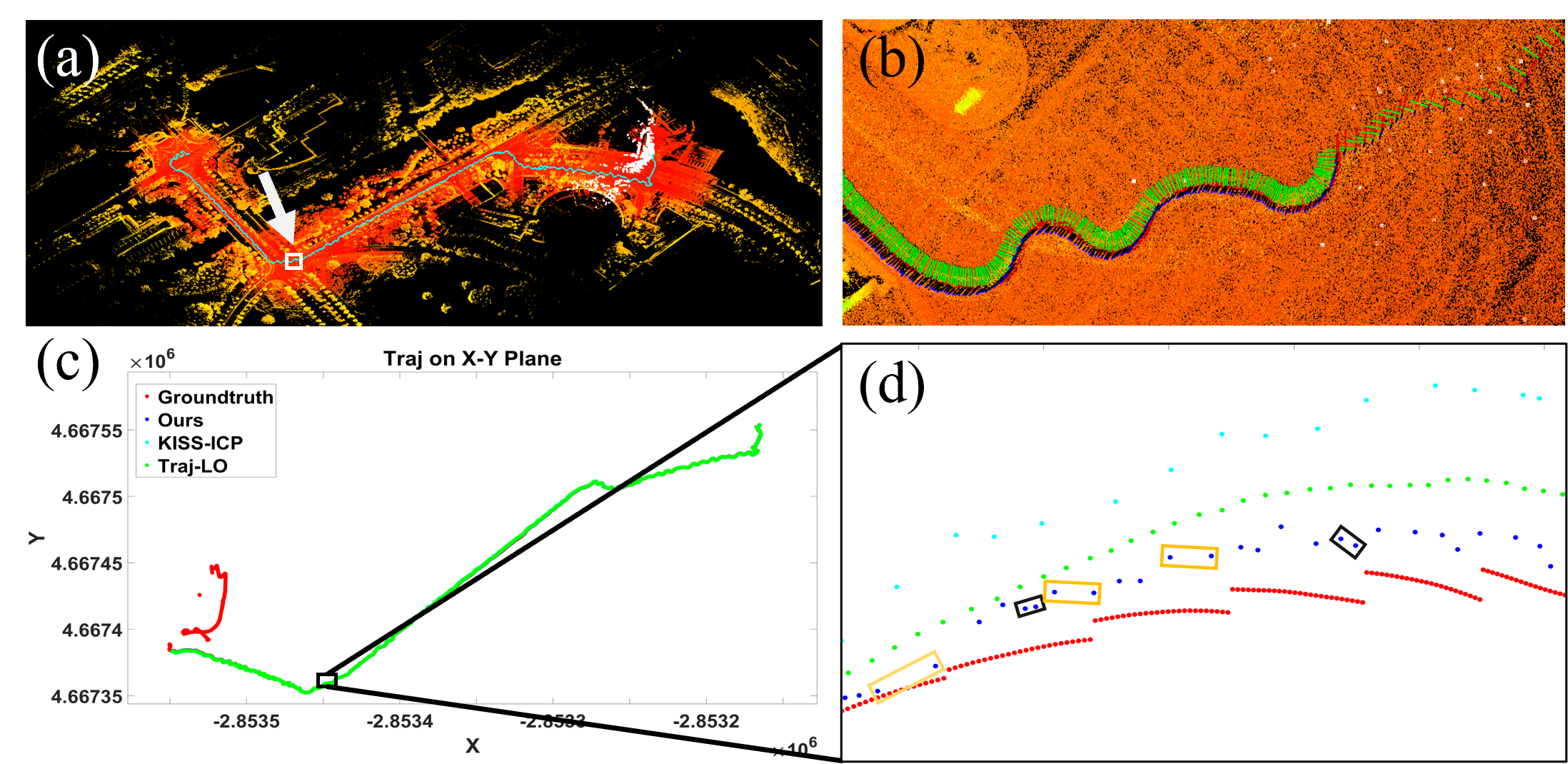}
	\caption{\small Results of LiDAR odometry on \textit{street\_07} sequence of M2DGR. \textbf{(a)} Our method, the {white} box marks the position where CT-ICP drifts. \textbf{(b)} {CT-ICP shows significant errors at the marked position in \textbf{(a)} }. \textbf{(c)} Trajectory results aligned with the ground truth. \textbf{(d)}  A zoomed-in view, where orange boxes indicate increased temporal intervals between control nodes due to degeneracy management, and black boxes indicate reduced intervals due to PCA-assessed motion state changes. }
	\label{fig:street7}
	\vspace{-0.2in}
\end{figure} 
We first validate our algorithm's effectiveness under vehicle conditions. The M2DGR dataset, collected by a unmanned ground vehicle (UGV). 
{We conduct tests in various environments and compute the Root Mean Squared Error (RMSE) of Absolute Trajectory Error (ATE) and {Relative Pose Error (RPE)} for the trajectories. 
	As shown in Table \ref{tab:m2dgr}, while the average improvement in accuracy on the UGV is modest (5\%, {calculated from ATE}), our method achieves a remarkable improvement (14\%, {calculated from ATE}) in the highly dynamic sequence (\textit{street\_07}). As seen in Fig.~\ref{fig:street7}(d), our method continuously adjusts the temporal intervals, resulting in a more accurate alignment with the ground truth.} 

\subsubsection{NTU VIRAL Dataset}
\begin{figure}[h]
	\centering
	\begin{subfigure}[b]{0.48\textwidth}
		\centering
		\includegraphics[width=0.48\textwidth]{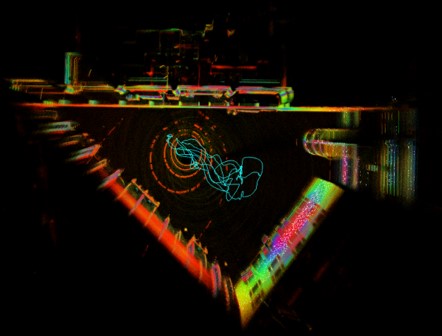}
		\includegraphics[width=0.48\textwidth]{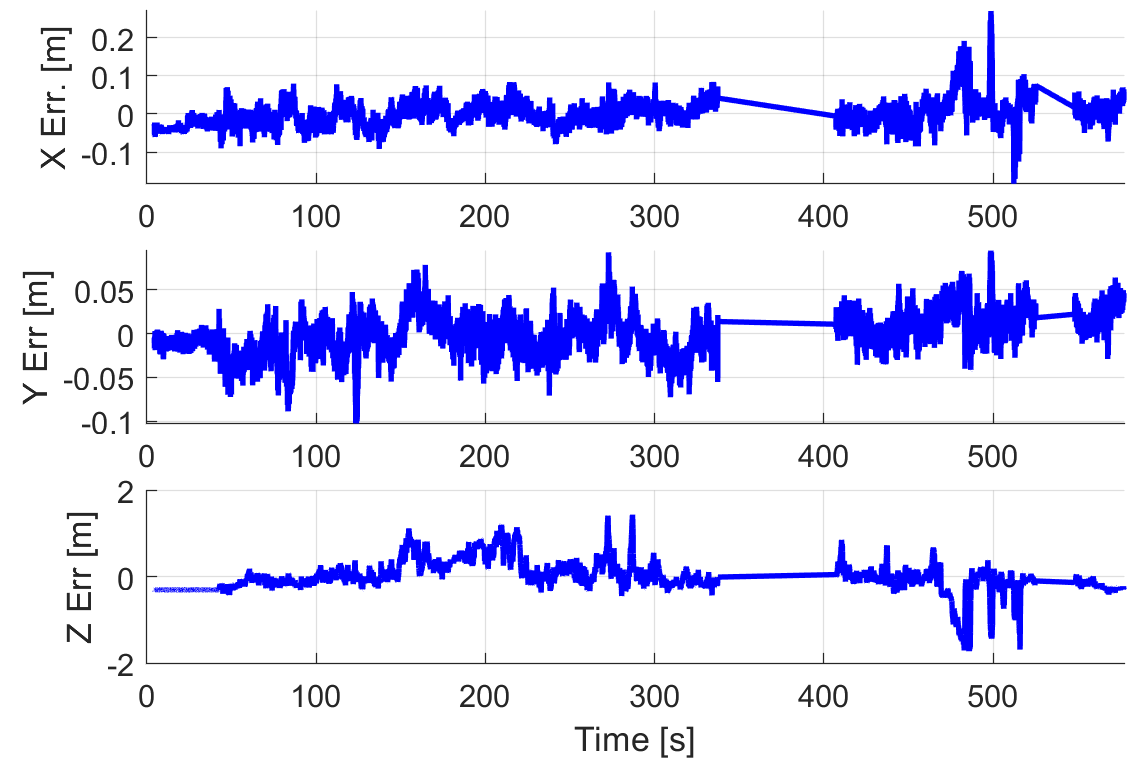}
		\caption{\textit{tnp\_01}}
		\label{fig:tnp_01_map}
	\end{subfigure}
	\hfill
	\begin{subfigure}[b]{0.48\textwidth}
		\centering
		\includegraphics[width=0.48\textwidth]{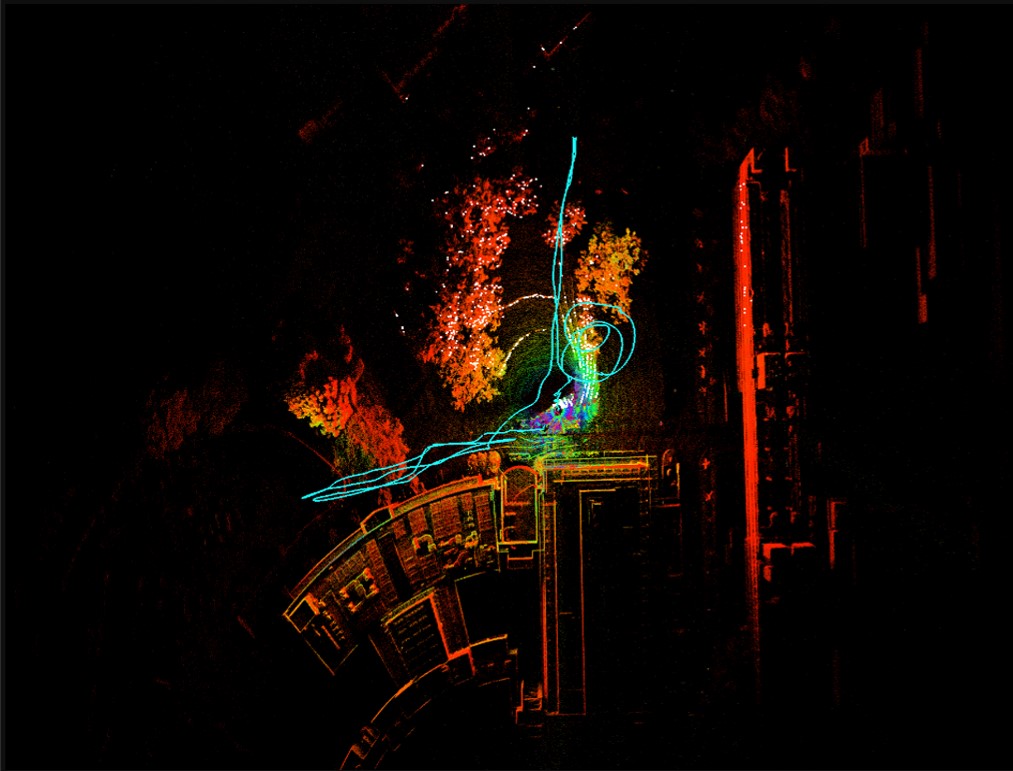}
		\includegraphics[width=0.48\textwidth]{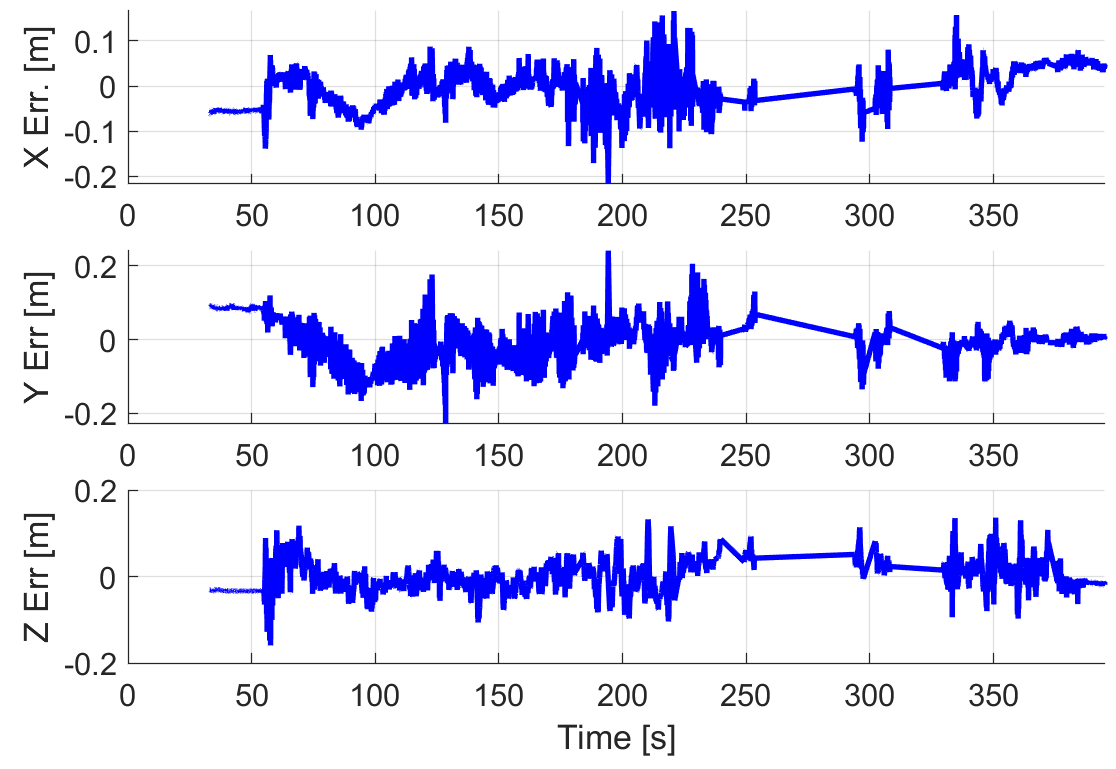}
		\caption{\textit{spms\_02}}
		\label{fig:spms_02_map}
	\end{subfigure}
	\caption{\small The environments and errors of our method on \textit{tnp\_01} and \textit{spms\_02} sequences. The Z-axis error is significantly greater than the X and Y axes, indicating that most trajectory errors stem from the Z-axis.} 
	\label{fig:tnp01_spms02}
	\vspace{-0.2in}
\end{figure}

\begin{figure*}[ht]
	\centering
	\includegraphics[width=\textwidth]{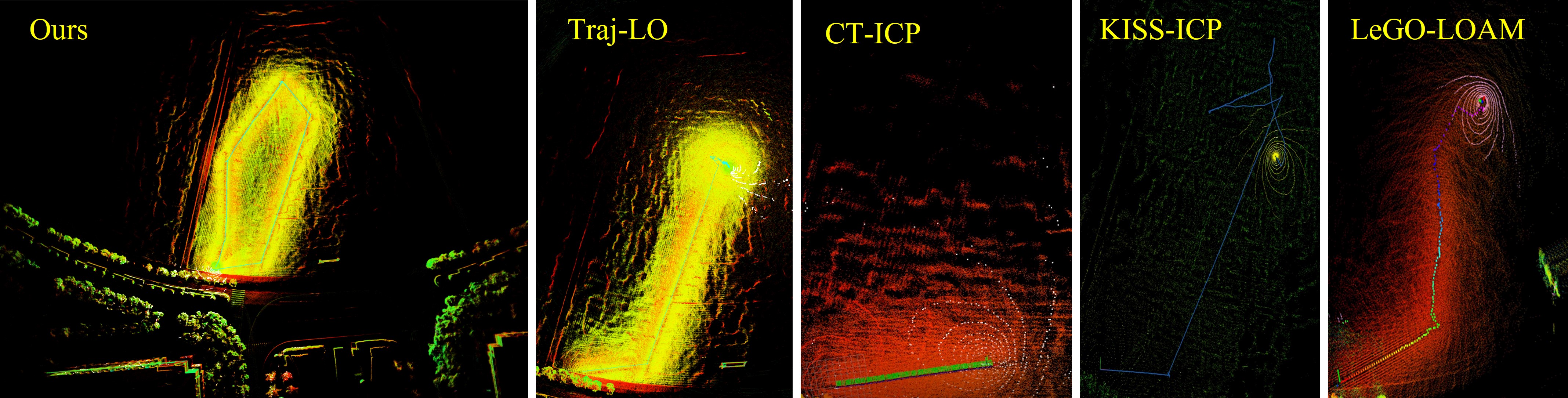}
	\caption{\small Results on \textit{hard} sequences. Our method is the only algorithm that successfully completed the \textit{hard} sequence among tested methods.  }
	\label{fig:hard_traj}
	\vspace{-0.2in}
\end{figure*}

The NTU VIRAL dataset, recorded by a UAV equipped with two Ouster LiDAR sensors (one horizontal, L1, and one vertical, L2), includes multiple sequences captured in various challenging indoor and outdoor conditions.
All algorithms are tested using the L1 LiDAR, and error evaluations are based on script provided by the dataset authors\footnote{https://ntu-aris.github.io/ntu\_viral\_dataset/evaluation\_tutorial.html}, {which computes the RMSE of ATE}. Since LeGO-LOAM is designed for ground robots, we employ another feature-based method F-LOAM~\cite{wang2021f} instead. 
{As shown in Table~\ref{tab:ntu_dataset}, our method achieves the highest accuracy in sequences \textit{eee}, \textit{nya}, \textit{sbs}, \textit{rtp}, \textit{spms}. Notably, in the \textit{spms\_02} sequence, the sparsity of environmental features (Fig.~\ref{fig:tnp01_spms02}(b)) caused most algorithms to diverge. In contrast, our method, thanks to its degeneracy management, merges subsequent cloud segments and ensures stable performance. More details can be seen in Section.~\ref{ablation study}.}

\begin{figure}[ht]
	\centering
	\includegraphics[width=0.48\textwidth]{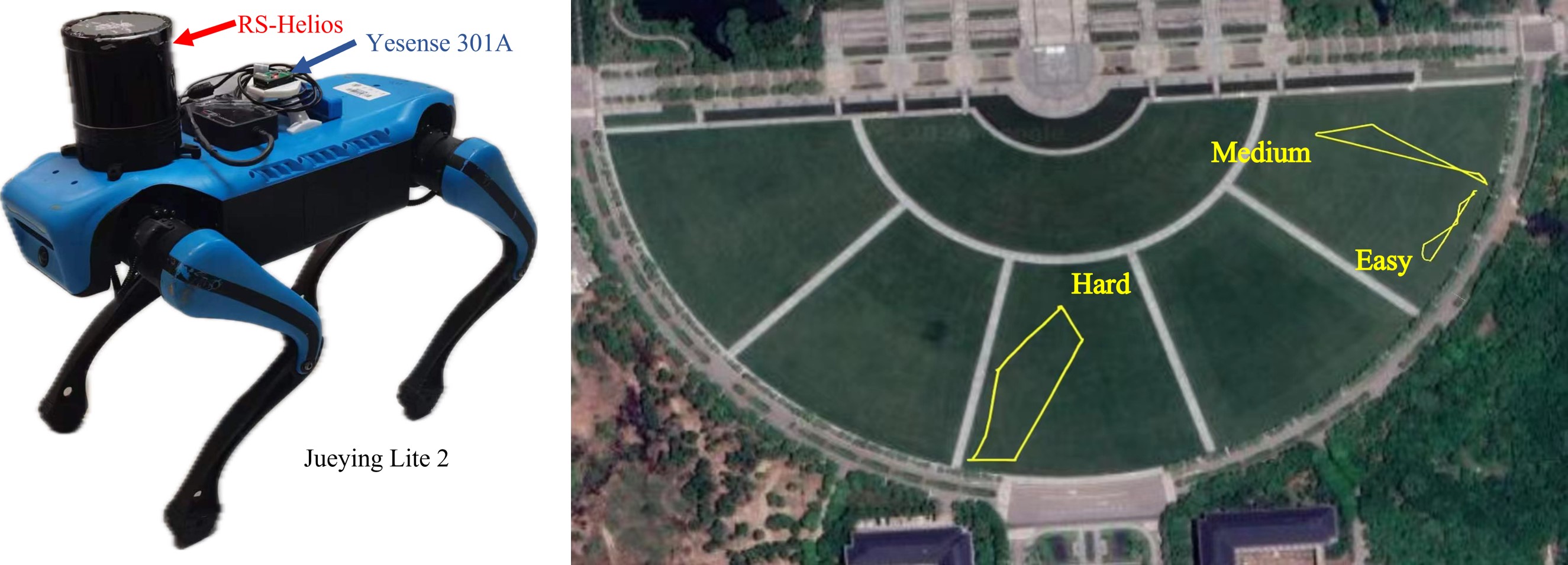}
	\caption{Jueying Lite2 and Jiulonghu Campus Dataset. }
	\label{fig:jiulonghu}
	\vspace{-0.2in}
\end{figure}

{In the \textit{tnp} sequence, the surrounding buildings result in insufficient constraints along the z-axis (Fig.~\ref{fig:tnp01_spms02}(a)). CT-ICP maintains lower ATE due to its broader neighborhood search range (27 closest voxels)\cite{zheng2024traj}. Despite using a smaller search range of 7 closest voxels (similar to Traj-LO), our method achieves high accuracy, ranking second only to CT-ICP. The degeneracy management module helps by decreasing the downsampling rate ($D_s$) and merging point clouds when z-axis constraints are insufficient, minimizing the effect.}

\subsubsection{Our Own Jiulonghu Campus Dataset}

In addition to public datasets, we validate the effectiveness of our method under challenging conditions. As shown in Fig.~\ref{fig:jiulonghu}, we collected challenging sequences using the quadruped robot \textit{Jueying Lite2}, which is equipped with a Robosense RS-Helios 32-line LiDAR, on the lawn of Southeast University's \textit{Jiulonghu} Campus. The sequences labeled \textit{easy}, \textit{medium}, and \textit{hard} increase in difficulty and distance from the lawn edge. In the easy sequence, the robot moves along the lawn's edge. In the medium sequence, the robot moves towards the center of the lawn but still has some trees and other obstacles providing LiDAR points. In the hard sequence, the robot is in the middle of the lawn, surrounded only by the ground and sparse edge points, leading to significant degeneracy. As shown in Table~\ref{tab:jiulonghu_dataset}, among the tested LiDAR-only odometry algorithms, ours is the only one that successfully completes all sequences.

\begin{table}[ht]
	\centering
	\caption{Jiulonghu Campus Dataset Results}
	\resizebox{0.49\textwidth}{!}{
		\begin{threeparttable}
			\begin{tabular}{c c c c c c c}
				\toprule
				\midrule
				Sequences &  LeGO-LOAM & KISS-ICP & CT-ICP & Traj-LO & Ours\\
				\midrule
				easy(127.86m) & \ding{55} & \checkmark & \ding{55} & \checkmark & \checkmark \\
				medium(92.44m) & \ding{55} & \ding{55} & \ding{55} & \checkmark & \checkmark \\
				hard(522.05m) & \ding{55} &\ding{55} & \ding{55} & \ding{55} & \checkmark \\
				\bottomrule
			\end{tabular}
			\begin{tablenotes}
				\small{\item[1] {'\ding{55}' denotes significant error, as defined in the same way as in Table~\ref{tab:m2dgr}}}
			\end{tablenotes}	
		\end{threeparttable}
		
	}
	\label{tab:jiulonghu_dataset}
	\vspace{-0.1in}
\end{table}
{Notably, in the \textit{hard} sequence, similar to the \textit{spms\_02} sequence from the NTU dataset, our method shows remarkable performance in environments with sparse features. Results are shown in Fig.~\ref{fig:hard_traj}.}

\subsection{Ablation Studies} \label{ablation study}

\begin{table}[ht]
	\centering
	\caption{RMSE of ATE (m) on Ablation Studies}
	\resizebox{0.48\textwidth}{!}{
		\begin{threeparttable}
			\begin{tabular}{c c c c c}
				\toprule
				\midrule
				\multirow{2}{*}{Approach} & \multicolumn{3}{c}{Newer College Dataset} & NTU VIRAL Dataset \\
				\cmidrule(r){2-4}
				\cmidrule(r){5-5}
				& 05\_Quad(479.0m) & 06\_Dynamic(97.2m) & 07\_Parkland(695.7m) & spms\_02(697.9m) \\
				\midrule
				KISS-ICP & 0.298 & \ding{55} & 0.260 & \ding{55} \\
				CT-ICP & \textbf{0.098} & 0.143 & 0.180 & \ding{55} \\
				Traj-LO & 0.334 & 0.146 & \ding{55} & \ding{55} \\
				\midrule
				Ours(None)& \ding{55} & 0.186 & \ding{55} & \ding{55}\\
				Ours(PCA) & \ding{55} & \textbf{0.124} & 0.131 & \ding{55}\\
				Ours(DM) & 0.175 & 0.184 & \ding{55} & 0.126\\
				Ours(PCA+DM) & 0.119 & \textbf{0.124} & \textbf{0.130} & \textbf{0.077}\\			
				\bottomrule
			\end{tabular}
			\begin{tablenotes}
				\small{\item[1] '\ding{55}' denotes significant error, as defined in the same way as in Table~\ref{tab:m2dgr}. \quad '(*m)' denotes the length of the sequence. 
					\item[2] The results of CT-ICP on Newer College Dataset are obtained from\cite{chen2023direct}.}
			\end{tablenotes}	
		\end{threeparttable}
		
	}
	\label{tab:AlationStudy}
	\vspace{-0.1in}
\end{table}

{To further validate the effectiveness of our method, ablation studies are conducted on the Newer College Dataset, with the results presented in Table~\ref{tab:AlationStudy}.	{Ours(None)} represents our proposed method with both the PCA pre-assessment module and degeneracy management disabled. {Ours(PCA)} and {Ours(DM)} refer to {Ours(None)} with the addition of either \textit{PCA} pre-assessment or the \textit{D}egeneracy \textit{M}anagement, respectively. {Ours(PCA+DM)} represents the full implementation of our proposed method.}

{The importance of the PCA pre-assessment in dynamic motion is demonstrated in the \textit{06\_Dynamic} and \textit{07\_Parkland} sequences. In the \textit{06\_Dynamic} sequence, characterized by aggressive motion with rotational speeds up to 3.5 rad/s,  the addition of PCA results in a 13\% accuracy improvement compared to the SOTA methods. In \textit{07\_ParkLand}, sudden motion changes cause failures in Ours(None) and Ours(DM), but the PCA module reduces the temporal interval, improving the algorithm's effectiveness.}

{Ablation experiments on the \textit{spms\_02} and \textit{05\_Quad} sequences highlight the robustness enhancement brought by the degeneracy management module. In the \textit{05\_Quad} sequence, Ours(None) and Ours(PCA) diverge due to the narrow starting space, while Ours(DM) adapts to the indoor environment with degeneracy management. Ours(PCA+DM) further improves outdoor accuracy with PCA pre-assessment. In the \textit{spms\_02} sequence, Ours(DM) integrates subsequent point cloud data to acquire sufficient observational constraints, ensuring the stability of the algorithm. }

\subsection{Real-Time Analysis}


\begin{table}[h]
	\centering
	\caption{The Pose Output Temporal Interval}
	\resizebox{0.48\textwidth}{!}{
			\begin{tabular}{c c c c c c c}
				\toprule
				\midrule
				Sequences &  \textit{street\_07} & \textit{spms\_02} & \textit{hard} \\
				\midrule
				Average Interval(ms) & 64.16 & 82.89 & 48.78 \\
				\bottomrule
			\end{tabular}
		}
		\label{tab:pose_hz}
		\vspace{-0.1in}
	\end{table}
	
	Our method dynamically adjusts the temporal interval between control nodes, resulting in a non-fixed pose output frequency. We evaluated the pose output frequency across sequences from three platforms. As depicted in the Table~\ref{tab:pose_hz}, the \textit{street\_07} sequence is a standard road scene with continuous aggressive turns, yielding an average pose output rate of 64.16ms. In the \textit{spms\_02} environment, degeneracy requires continuous cloud segment fusion, resulting in a lower output rate (82.89ms). In contrast, the \textit{hard} sequence from the \textit{Jiulonghu} Campus Dataset, despite continuous cloud segment fusion, has sparser environmental features leading to a higher output rate (48.78ms). The overall pose output rate of our method falls between 10Hz and 20Hz.

\section{Conclusion}
	We introduce an adaptive continuous-time LiDAR odometry that dynamically adjusts the temporal intervals between control nodes based on motion dynamics and environmental features, offering significant advantages in handling more aggressive motions and open environments. The degeneracy management approach successfully addresses the degeneracy problems arising from point cloud segmentation in existing continuous-time methods. Experiments on various platforms demonstrate that our method achieves accuracy comparable to current SOTA methods and outperforms them in feature-sparse environments. Future work will focus on continuous-time loop detection techniques to enhance the performance of continuous-time methods in long-duration navigation scenarios.

\bibliographystyle{ieeetr}
\bibliography{REFERENCES}

\end{document}